\documentclass{article} 
\usepackage{nips12submit_e,times}
\usepackage{graphicx}
\usepackage{subfigure}
\usepackage{caption}
\usepackage{booktabs}
\usepackage{multirow}
\usepackage{amsmath}
\usepackage[sort]{natbib}
\usepackage{url}

\newcommand{\enotesoff}{\long\gdef\enote##1##2{}}
\newcommand{\enoteson}{\long\gdef\enote##1##2{{\bf
\hspace{1cm}$<<<$ ##1: ##2 $>>>$\hspace{1cm}}}}

\enoteson
\enotesoff

\title{Cutting Recursive Autoencoder Trees}

\author{
Christian Scheible \\
Institute for Natural Language Processing \\
University of Stuttgart, Germany \\
\texttt{scheibcn@ims.uni-stuttgart.de} 
\AND
Hinrich Sch\"utze \\
Center for Information and Language Processing \\
University of Munich, Germany \\
}

\nipsfinalcopy 

\begin{document}

\maketitle

\begin{abstract}
Deep Learning models enjoy considerable success in Natural
Language Processing. While deep architectures produce useful
representations that lead to improvements in various tasks,
they are often difficult to interpret. This makes the
analysis of learned structures particularly difficult. In
this paper, we
rely on empirical tests to see whether a
particular structure makes sense. We present
an analysis of 
the Semi-Supervised
Recursive Autoencoder,
a well-known model that produces
structural representations of text. We show that for certain tasks, the
structure of the autoencoder can be significantly reduced
without loss of classification accuracy
and we evaluate the produced structures using human
judgment.

\end{abstract}

\section{Introduction}

Deep Learning (DL) approaches are gaining more and more
attention in Natural Language Processing. Many NLP tasks
have been addressed:
syntactic parsing \citep{dlparsing}, semantic role labeling
\citep{scratch}, machine translation \citep{dlmt}, and
document classification \citep{glorot}.

One important issue in applying DL lies in the structure of
language. DL originated in image processing where the
neighborhood of two entities, e.g. pixels, has a
straightforward interpretation through spatial relations. In
NLP, neighborhood relations are not always interpretable
easily and don't necessarily translate into a semantic
relationship. Often, long distance dependencies hinder the
use of locally sensitive models.

A frequently cited DL model for NLP tasks is a model
introduced by \citep{socher2011}, the Semi-Supervised
Recursive Autoencoder (RAE). This model aims for a
compositional approach, representing each word as a vector
and recursively applying an autoencoder function to unify
pairs of nodes, yielding a tree. One appealing property of
RAEs is that the resulting structure lends itself to a
syntactic or even semantic interpretation. However, so far
no in-depth analysis of RAEs in terms of these structures
has been performed.

Our main interest in this paper is to analyze the behavior
of RAEs. We will investigate the following key questions:
(i) Are the structures produced by RAEs interpretable
syntactically or semantically? (ii) Can the structures be
simplified?

We will analyze these issues empirically in a sentiment
classification task that RAE has previously been used for
\citep{socher2011}. We introduce two methods for
analysis. First, we try to simplify the RAE structures
automatically and evaluate the resulting models on a
classification task. Second, we let humans rate the
structures according to syntactic and semantic criteria.

In Section~\ref{sec:method}, we describe RAEs, particularly
highlighting some details regarding implementation. We then
turn to different ways of structural simplification in
Section~\ref{sec:simplify}. Section~\ref{sec:task} 
introduces the task we use for
evaluation. Section~\ref{sec:analysis} contains error
analysis of RAEs conducted by human annotators. In
Section~\ref{sec:experiments} we carry out the experiments
on structural simplifications.

\section{Semi-Supervised Recursive Autoencoders}
\label{sec:method}

The central model in this paper is the Semi-Supervised
Recursive Autoencoder (RAE) \citep{socher2011}. This section
describes this model  and
discusses some important implementation details.
 
The RAE is a structural model. It recursively applies a
neural network, the autoencoder, to construct a tree
structure over the words in a sentence. Each word is
represented as a vector which is independent of the context
in which the word occurs. In addition to the usual autoencoder 
objective of reconstruction, the representation at
each node is also used to predict the class of the whole 
sentence by applying a softmax neural network to the nodes.

The basic representations of words are randomly initialized
vectors of dimensionality $h$, stored in a matrix $W$ where
every row represents one word. This representation is
enhanced using an embedding matrix $L$ which is optimized
during training. The final representation of a word indexed
by $n$ is obtained by $W_n + L_n$. This representation
serves as the base representation for tree
construction.\footnote{Some of these details are best described in
the RAE implementation available at
\url{http://www.socher.org}.} First, we assume the words to be the 
leaf nodes of the tree. Trees are then constructed by
iteratively joining two adjacent nodes using an
autoencoder which consists of two layers. The encoding layer 
takes two nodes $n_1$ and $n_2$ and outputs a combined 
representation $r$:

\begin{equation*}
r = f(A_1 [n_1;n_2] + b_1)
\end{equation*}

Subsequently, the reconstruction layer tries to reproduce the
original inputs:

\begin{equation*}
[n_1, n_2] = A_2 \, r + b_2
\end{equation*}

$f$ is a non-linear function. $A_x$ is the weight matrix,
$b_x$ the bias for the respective layer. Note that the
dimensionality of $r$ needs to be the same as of $n_1$ and
$n_2$ so that the autoencoder can be applied recursively.

The resulting output of the autoencoder (again of
dimensionality $h$) serves as the representation of a new
node that has the two encoded nodes as its children. The
combination operation is carried out greedily by
autoencoding the pair of nodes first that has minimal
reconstruction error $E_\textrm{rec}$.

Each node output is then used to predict the sentence label
individually using a softmax layer:

\begin{equation*}
c = \textrm{softmax}(A_l \, r),
\end{equation*}

where $A_l$ is a weight matrix and $r$ the representation of
a node.

The representations have an influence on both reconstruction
and classification. Therefore, there are two objectives to
be minimized: the reconstruction error $E_\textrm{rec}$ that
specifies how well the resulting node represents the two
children, and a classification error $E_\textrm{cl}$ that
measures how well the correct label of the sentence can be
predicted from the information at the node. $E_\textrm{rec}$
is the Euclidean distance between the original and
reconstructed nodes. $E_\textrm{cl}$ is the cross-entropy
error between the correct label and the output of a softmax
layer that is trained. In particular, the embedding matrix
$L$ is optimized by calculating the classification errors
over all words in the training set. For batch optimization
arithmetic means of the errors $\bar E_\textrm{rec}$ and
$\bar E_\textrm{cl}$ and the corresponding gradients are
calculated over all nodes. Higher-order nodes are penalized
in favor of leaf nodes with a factor
$\beta$. $E_\textrm{rec}$ and $\bar E_\textrm{cl}$ are added
with weight $\alpha, 1-\alpha$. The model parameters are
optimized with L-BFGS.

After the autoencoder is trained, a feature extraction step
follows.  Following \cite{socher2011}, we performed this
step differently from the way the autoencoder is optimized.
First, we recursively apply the aforementioned greedy
autoencoder to build a tree for each sentence. In contrast
to the RAE training, where the errors of individual nodes
are averaged, we first calculate the arithmetic mean of all
node features to get a single feature for the tree.  Let
$f_1 \ldots +f_n$ be the features of the $n$
nodes in a tree. Feature extraction returns $\bar f =
\frac{\sum_{i}f_i}{n} $ as the tree's
representation. Finally, a softmax neural network is trained
on this representation. Taking the mean of all nodes
resembles convolution operations which have been
successfully applied in NLP previously, e.g. by
\cite{scratch} who calculate the maximum over the dimensions
of their representation vectors.  The original
implementation \citep{socher2011} also uses the top feature
separately. We leave this feature out in our experiments as
it did not improve the results significantly.

The RAE has several parameters that need to be set. First, the
weighting of reconstruction and classification error
$\alpha$. Second, the penalty of higher-order node errors
versus leaf-level errors $\beta$. In addition, all activation matrices 
vectors are regularized with the $L_2$ norm. We adopt the parameter
settings used in \citep{socher2011}.

\section{Structural simplifications}
\label{sec:simplify}

In a complex model like the RAE it is difficult to see which
components are responsible for the results it achieves. In
order to analyze these structures, we try to simplify them
automatically. Evaluating the model on a task will then show
us whether the structure omitted made a contribution. In the
following sections we will present three ways of structural
simplification which we will apply to the RAE trees.

\subsection{Tree level cuts}

As a first, simple approach to determine the influence of
higher level nodes in the tree, we simply remove nodes from
the representation. One straightforward operation that
achieves this is a level cut. 

We count levels starting at the leaves, the
basic units for the RAE. All terminal nodes $t$ are defined
to have level $l(t)=1$, and each non-terminal $n$ with
children $\langle c_1, c_2\rangle$ has the level $l(n) =
\textrm{max}(l(c_1), l(c_2))$. In practice, we compute the
full tree and then prune away all nodes that have a level
higher than $l_\textrm{max}$. We call this a (tree) \emph{level cut}.

\subsection{Subtree selection}

Another approach to simplification follows from the idea
that not every word is important for sentiment
classification, but rather that there is a region in the
sentence that is sufficient to recognize sentiment
(cf. \citep{subtrees}). A good example is the tree in
Figure~\ref{fig:sentence2}. In order to predict the correct
sentiment of the sentence, it is sufficient to analyze the
second clause (``it never took off and always seemed
static"). This way of
simplification is orthogonal to level cuts. Level cuts
reduce the amount of structure induced by the autoencoder
but keep the complete input. Subtree selection reduces the
input, but it makes use of and is based on the full tree
representation.

In order to select a region, we greedily select a central
word: We apply the softmax function of the autoencoder to
each word in the sequence and pick the one with the lowest
$E_\textrm{cl}$ as the central word. For training examples,
we compute the error for the gold class. For testing
examples, we compute the score for all classes and select
the word with the overall minimal error. Starting from this
point, we select the largest subtree of the tree produced by
the RAE whose top node $n$ has a level of at most 
$l_\textrm{max}$.

\subsection{Window selection}

Related to this is the window approach. Here, we again
identify a central word as shown before and take the
representations of all words within a window $w$ to either
side as input to the classifier, where the central word is 
the left- or rightmost word, respectively. All other words are
ignored. For example $w=3$ means that we take the two
words to the left and to the right of the central word
and drop everything else; and $w=1$ only uses the
central word and no context. In this approach, no tree
structures are used, only the embeddings.

\section{Task}
\label{sec:task}

We 
use the same task and data set for our RAE investigation
as \citep{socher2011}:
sentiment classification for
the sentence polarity dataset by \cite{panglee}. It
contains 10{,}662 sentences from movie reviews that were
manually labeled as expressing positive or negative
sentiment towards the movie. We use the implementation 
provided at \url{http://www.socher.org}.
We set the hidden layer dimensionality
to $h=50$. All experiments are carried out on a random 90/10
training-testing split. Accuracy is used as the evaluation
measure since the class distribution in the dataset is
balanced.

\section{Error Analysis}
\label{sec:analysis}

Error analysis proves to be difficult for automatically
generated representations. In general, the dimensions
produced by autoencoding in NLP applications cannot be
interpreted easily.  Therefore, we resort to empirical
evaluation in the context of our task. In this section, we
will provide analyses conducted by human annotators of two
properties of the trees: syntactic and semantic
compositionality.

\begin{figure}
        \centering
        
         \subfigure[Sentence 1: \emph{not a bad journey at
             all .}]{ \includegraphics[width=.6\textwidth]
           {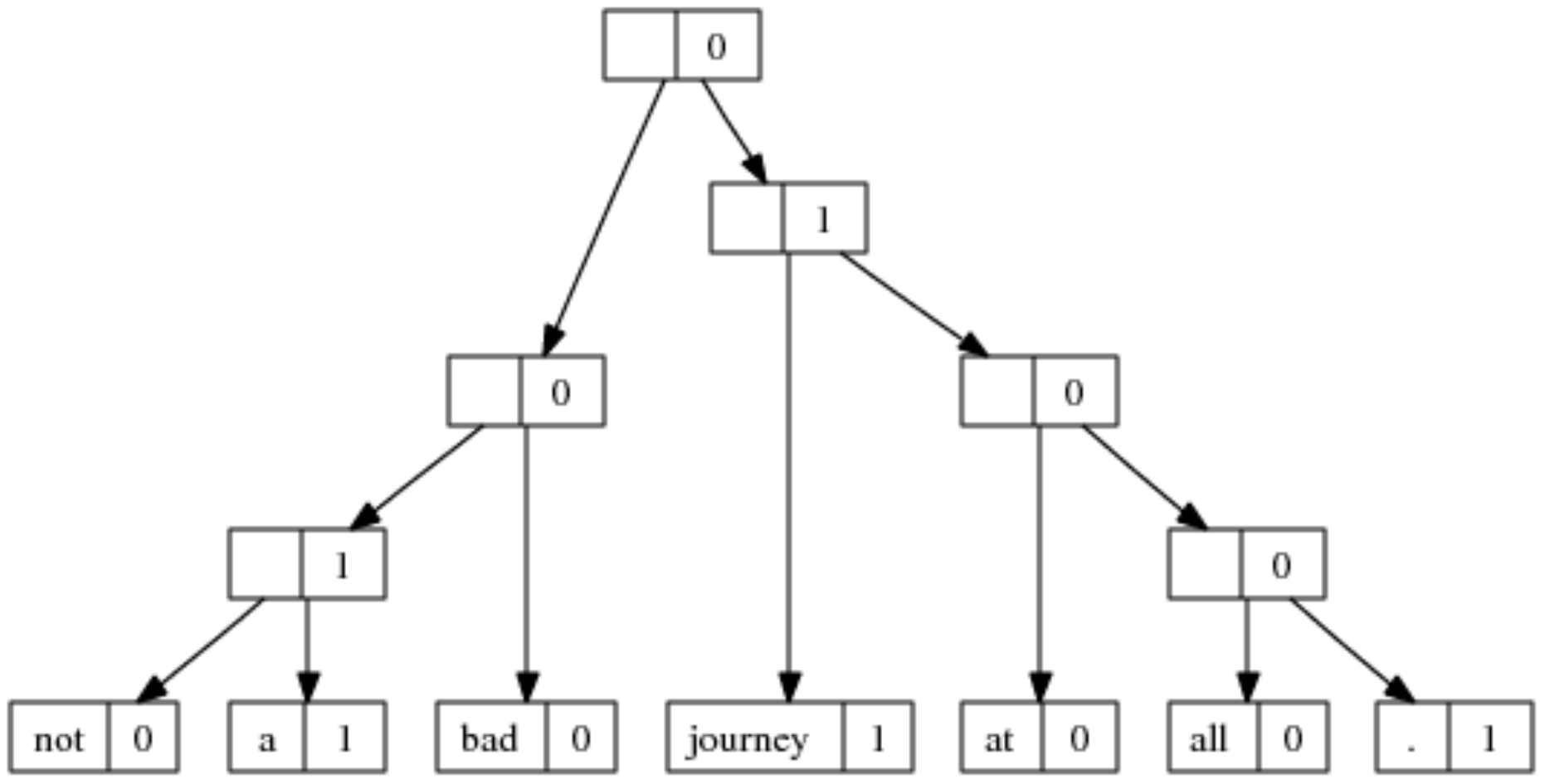}
   \label{fig:sentence1}
 }

 \subfigure[Sentence 2: \emph{though everything might be
     literate and smart , it never took off and always
     seemed static .}]{ \includegraphics[trim=15cm 0cm 0cm
     4cm, clip=true, width=\textwidth] {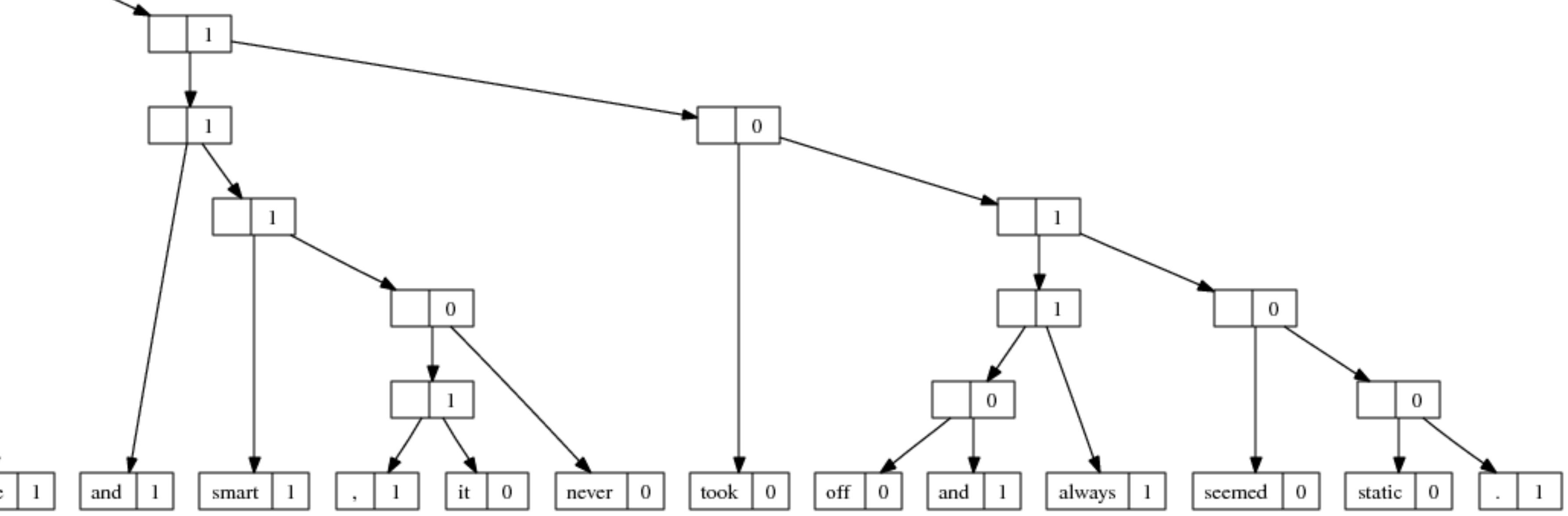}
   \label{fig:sentence2}
 }

\subfigure[Sentence 3: \emph{the *UNKNOWN* elaborate
    continuation of `` the lord of the rings " trilogy is so
    huge that a column of words can not adequately describe
    *UNKNOWN* peter *UNKNOWN* 's expanded vision of
    *UNKNOWN* . r . r . *UNKNOWN* 's *UNKNOWN* .}]{
  \includegraphics[trim=50cm 0cm 50cm 6cm, clip=true,
    width=\textwidth] {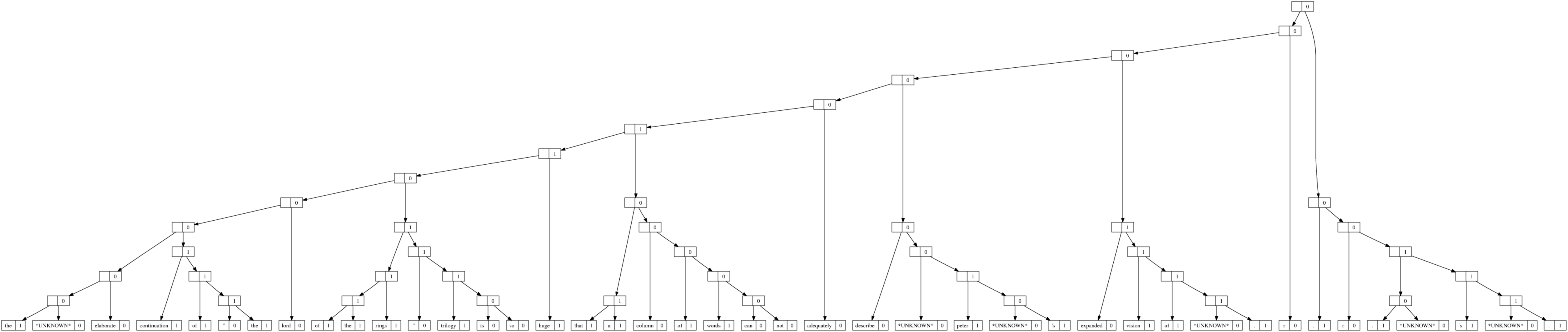}
   \label{fig:sentence3}
 }

        \caption{Example trees. Leaf nodes contain the word
          that is encoded by them. All nodes contain the
          sentiment predicted using softmax, 1 being
          positive and 0 negative sentiment. 
We follow \citep{socher2011}
and replace
words 
for which no embedding is available 
by ``*UNKNOWN*''.}
        \label{fig:sentences}
\end{figure}

\subsection{Syntactic Coherence}
Naturally, it is tempting to think of RAE tree structures in
terms of syntactic analysis. In this section, we will show
that there is a large divergence between traditional
syntactic trees as most theories of grammar would posit and
the trees produced by RAE.
We analyze two phenomena:
coordinating conjunctions and negation. While coordinations
are notoriously difficult even in supervised problems,
negations are less problematic \citep{mcdonald, collins}. We
asked 2 humans to judge whether the parse of 10 randomly
selected examples for each of the phenomena was correct with
respect to the phenomenon.  None of the 20 parses (10
for each of the two phenomena) were unanimously determined to be
correct by the human judges.

The example trees in Figure~\ref{fig:sentences} illustrate
these results. We will first take a look at examples for
negation. In sentence~\ref{fig:sentence1}, the autoencoder
used \emph{not} at a low level in the tree, constituting a
modification of \emph{a}. Their joint representation is
itself joined with \emph{bad}. The correct analysis would
use \emph{not} as a modifier to a joint structure \emph{a
  bad journey} where \emph{bad} and \emph{journey} are
combined directly. Sentences~\ref{fig:sentence2}
and~\ref{fig:sentence3} represent cases where the
autoencoder introduced long distances between the negating
and the negated phrase (\emph{never} to \emph{took off} and
\emph{not} to \emph{describe}).

We now turn to coordinations. In
sentence~\ref{fig:sentence2}, we find an instance of a
coordination of two clauses. The clause \emph{always seemed
  static} should receive a joint analysis and should then be
modified by \emph{and}. Instead, \emph{and} is put in a
subtree containing two words from each of the coordinated
clauses.

The underlying reason for the resulting structures may
actually arise through the property of greediness. As RAEs
are trained greedily by joining the least error-prone
combinations first, pairings of frequent words are common.
For example, in sentence~\ref{fig:sentence1}, frequent words
are joined first (\emph{not} to \emph{a}, and \emph{all} to
\emph{.}). The most uncommon words are added last
(\emph{bad} and \emph{journey}). In around 75\% of all
occurrences, periods are adjoined directly to their
neighbors, which is not desirable from a syntactic point of
view.

\subsection{Semantic Coherence}
We will now analyze the behavior of RAEs from a semantic
point of view. Sentiment analysis, our example task, relies
heavily on semantic composition.

Usually, sentiment is only carried by a small number of
expressions, for example adjectives like \emph{great} or
\emph{awful} in combination with their close syntactic
environment \citep{subtrees}. The sentiment of an expression
can then be modified by applying an intensifying or
reversing construction. A simple check for whether RAEs are
able to learn compositionality is to check the produced
trees for instances of these modifiers and see whether they
behave as expected.

Intensifiers such as \emph{very} or \emph{little} are
difficult to analyze as there is no straightforwardly
quantifiable result that one would expect to occur,
especially in the case of discrete labels as opposed to
continuous sentiment scores. Therefore, we only consider
reversers in this analysis. There is no consensus as to
which words constitute the set of reversers. Often,
reversing properties are context-dependent
\citep{polanyi2006}. We therefore picked a small set of
reversers with general applicability: \emph{not}/\emph{n't},
\emph{no}, and \emph{never}.

To check whether reversal occurs in a tree, we first
calculate the classification decision by evaluating the
softmax decision function at each node. We search the trees
for occurrences of any reverser and check whether its
sibling and its parent node are assigned opposite classes,
which should be the case if the reverser was correctly
applied. We find that reversal happens only in around 31\%
of all reverser occurrences.

Of course, since reversing is context-dependent, we need
human input to verify this result. From all trees containing
reversers, we randomly selected 
3 trees in which the reverser reverses sentiment and 7 trees
in which the reverser does not reverse sentiment (the goal
being to mimick the 31\% rate of reversals).
 We asked 2 human judges to check
the examples for correct reversal. The judges unanimously
found that only 3 out of 10 candidates are behaving
correctly, confirming that reversal is very likely not a
property captured correctly by the RAE model.

We again turn to Figure~\ref{fig:sentences} for examples of
errors. In sentence~\ref{fig:sentence1}, \emph{not} does not
reverse the polarity of \emph{a} -- which is
correct. However, modifying \emph{bad} with the resulting
structure still does not reverse. The polarity of the whole
sentence seems to be determined by the polarity of
\emph{journey} which gets reversed at the top node, leading
to misclassification. In sentence~\ref{fig:sentence2},
\emph{never} should reverse \emph{took off}, yielding a
negative sentiment overall. However at the point where the
two phrases are joined, the topmost node depicted, positive
sentiment is predicted.

We reiterate that
the effect of reversers is quite complex and in
many contexts --
e.g., ``not
awesome, but pretty good'' --
they do not simply reverse sentiment.
However, the examples seem to show that the syntactic and
semantic role of reversers
is not modeled well in thoses cases where they act as simple reversers.

\section{Automatic structural simplification}
\label{sec:experiments}

In the previous section we showed that 
the structures produced by RAEs cannot be easily interpreted
in terms of traditional linguistic categories from syntax
and semantics.
We will now turn to empirically evaluating the
contributions of these structures in a practical
classification task.

\subsection{Level cuts during feature extraction}

\begin{table}[t!]
\begin{center}
{\raggedleft
\begin{tabular}{p{1cm} p{1.3cm}p{1.3cm}p{1.3cm}p{1.3cm}p{1.3cm}}
\toprule
\multirow{2}{*}{$l_{\textrm{max}}/w$}  & \multirow{2}{*}{extract} & train+ extract & no-embed & \multirow{2}{*}{subtree} & \multirow{2}{*}{window}  \\
\midrule
1        & 77.30 & 77.67 & 58.07 & 25.33 & 25.33 \\
2        & 75.98 & 68.95 & 62.89 & 25.33 & 30.96 \\
3        & 75.61 & 54.60 & 66.98 & 26.17 & 73.55 \\
5        & 74.86 & 52.60 & 69.14 & 27.67 & 74.20 \\
7        & 76.08 & 54.04 & 72.20 & 55.25 & 74.86 \\
10       & 76.17 & 62.02 & 74.11 & 71.58 & 75.61 \\
15       & 76.75 & 63.23 & 77.39 & 77.02 & 77.20 \\
20       & 76.75 & 63.60 & 77.49 & 76.74 & 77.20 \\
$\infty$ & 76.75 & 77.24 & 77.49 & 77.20 & 77.21 \\
\bottomrule
\end{tabular}
}
\end{center}
\caption{Accuracy using different structure reduction techniques.}
\label{tab:results}
\end{table}

In the first experiment, we train the autoencoders to
produce full trees and only apply level cuts in the feature
extraction. We report accuracies in Table~\ref{tab:results},
column \emph{extract} for different values of
$l_\textrm{max}$. The trees produced by the RAE on the whole
dataset have a mean height of 10 and a maximum height of 23.

First note that the best accuracy is achieved by cutting
directly above the leaves, i.e. using no trees at
all. Increasing the maximum level lowers accuracy
significantly; e.g. using $l_\textrm{max} = 5$, we lose over
2 percentage points compared to $l_\textrm{max} = 0$. Only
when $l_\textrm{max}$ is increased further, accuracy
recovers. This result suggests that the leaf representations
-- the embeddings -- carry great weight in classification.

In order to demonstrate the significance of the embeddings,
we again train full RAE trees but resort to
the random representations during feature extraction. 
The results of this run are
shown in Table~\ref{tab:results}, column \emph{noembed}.

Naturally, using random representations only we achieve low
accuracy. Note that the results are still over chance level,
an effect which may be caused by the random representations
being similar to low-dimensional random indexing
\citep{randomindexing}. Nevertheless, using higher-level tree
representations successively increases accuracy to a similar
level as observed in the previous experiment.

Our interpretation of this experiment is that while the
trees seem to be able to create useful representations of
the underlying words, these representations are redundant
with respect to the embeddings. Combining both does not lead
to improved classification accuracy.

\subsection{Level cuts during training}

One could argue that using a fully-trained RAE to extract
pruned trees is unfair since the model is still able to use
the full information induced during training. Taking the
level cut approach one step further, we also cut the trees
during training, using the same maximum level. Column
\emph{train+extract} in Table~\ref{tab:results} shows the
results for this experiment.

First, we observe that we get a well-performing model if the
maximum level is 1 in both RAE training and feature
extraction. This is not surprising as we are not using the
node-combining part of the RAE at all which makes this
particular model equal to the one with maximum level 1 in
the previous experiment. Next, we can see that accuracy
drops quickly as we introduce more levels and only recovers
after raising the threshold to $\infty$, using full trees. A
possible explanation for this phenomenon is that when
enforcing low levels there are also fewer training instances
for the RAE and thus the resulting models are worse. Another
possibility is that when full trees are constructed, all
applications of the RAE depend on each other since errors
are propagated through the structure. Thus, inconsistencies
should be optimized away. However, there are fewer
inconsistencies in lower-level cuts since the resulting
subtrees are likely to be disconnected.

These experiments show that the best accuracy
is achieved by a model that does not use the tree structures
at all. Our conclusion from this evidence is that the
strength of the RAE lies in the embeddings, not in the
induced tree structure.

\subsection{Subtree selection}

We now turn to subtree selection. As stated previously,
sentiment is a local phenomenon, so it might be sufficient
to use part of a sentence to classify the
data. Table~\ref{tab:results}, column \emph{sub} shows the
results for this experiment.

We observe low accuracies for low $l_{\textrm{max}}$. Only
when using large contexts (recall that the maximum number of
levels is around 23), the results become competitive. From
this experiment, it is not clear whether the height of the
trees or the size of the contexts -- which grows with the
height -- is responsible for the gain. We will investigate this
issue in the following section.

\subsection{Window selection}

As a last experiment, we concentrate on the embeddings as
they seem to be sufficient to achieve high accuracies on
this task. This will also show whether subtrees or embedded
words were responsible for the improvements with increasing
tree height in the previous section.

We vary the window size $w$ starting from 0 which is only
the word itself, to $\infty$ which is the maximum number of
words. The data has a mean sentence length of around 21
words and a maximum sentence length of 63 words. Results are
shown in Table~\ref{tab:results}, column \emph{win}.

While a small window size (0 or 1) produces bad results --
which might be an effect of choosing the wrong most
confident word resulting in strong overfitting -- vast
improvements are visible soon. Taking a window of 15 words
(in each direction) is sufficient. There are around 16\% of
the data that have more than 31 words, so it seems that for
sentences of that length, there are no context effects that
the model can exploit.

\subsection{Discussion}

We presented multiple experiments in which we simplified the
RAE tree structures. All of these experiments point towards
the embedding having the most influence on the end
result. If embeddings are not used, accuracy drops almost to
chance level. Using full trees and no embeddings seem to
have the same effect as using only embeddings. However,
using both representations together does not yield any
improvement. This suggests that there is a large overlap
between what the trees model and what the embeddings model.

\section{Conclusion}

In this paper, we conducted two different experiments concerning the
structures learned and generated by Semi-Supervised
Recursive Autoencoders. 

First, we automatically reduced the structure in different
ways and showed that on our sentiment analysis task the
embedded words were sufficient to achieve state-of-the-art
accuracy. Our experiments on window selection suggest that a
structure as simple as a well-chosen subset of the words in
a sentence produces a good model.

In a human evaluation, we 
showed that 
there is no simple way to interpret
the structures produced by RAEs in terms of traditional 
linguistic categories of syntax and semantics.

Overall, we conclude that structural simplifications are
possible at least for a sentiment analysis task.

\subsubsection*{Acknowledgements}
This work was supported by the Deutsche Forschungsgemeinschaft through the Sonderforschungsbereich 732.



\bibliographystyle{elsart-harv}
\bibliography{iclr2013}

\begin{thebibliography}{11}
\expandafter\ifx\csname natexlab\endcsname\relax\def\natexlab#1{#1}\fi
\expandafter\ifx\csname url\endcsname\relax
  \def\url#1{\texttt{#1}}\fi
\expandafter\ifx\csname urlprefix\endcsname\relax\def\urlprefix{URL }\fi

\bibitem[{Collins(1999)}]{collins}
Collins, M., 1999. Head-driven statistical models for natural language parsing.
  Ph.D. thesis, University of Pennsylvania.

\bibitem[{Collobert et~al.(2011)Collobert, Weston, Bottou, Karlen, Kavukcuoglu,
  and Kuksa}]{scratch}
Collobert, R., Weston, J., Bottou, L., Karlen, M., Kavukcuoglu, K., Kuksa, P.,
  Nov. 2011. Natural language processing (almost) from scratch. J. Mach. Learn.
  Res. 999888, 2493--2537.

\bibitem[{Deselaers et~al.(2009)Deselaers, Hasan, Bender, and Ney}]{dlmt}
Deselaers, T., Hasan, S., Bender, O., Ney, H., 2009. A deep learning approach
  to machine transliteration. In: Proceedings of the Fourth Workshop on
  Statistical Machine Translation. StatMT '09. Association for Computational
  Linguistics, Stroudsburg, PA, USA, pp. 233--241.

\bibitem[{Glorot et~al.(2011)Glorot, Bordes, and Bengio}]{glorot}
Glorot, X., Bordes, A., Bengio, Y., June 2011. Domain adaptation for
  large-scale sentiment classification: A deep learning approach. In: Getoor,
  L., Scheffer, T. (Eds.), Proceedings of the 28th International Conference on
  Machine Learning (ICML-11). ICML '11. ACM, New York, NY, USA, pp. 513--520.

\bibitem[{Kanerva et~al.(2000)Kanerva, Kristoferson, and
  Holst}]{randomindexing}
Kanerva, P., Kristoferson, J., Holst, A., 2000. Random indexing of text samples
  for latent semantic analysis. In: In Proceedings of the 22nd Annual
  Conference of the Cognitive Science Society. Erlbaum, pp. 103--6.

\bibitem[{McDonald(2006)}]{mcdonald}
McDonald, R., 2006. Discriminative learning and spanning tree algorithms for
  dependency parsing. Ph.D. thesis, University of Pennsylvania.

\bibitem[{Pang and Lee(2005)}]{panglee}
Pang, B., Lee, L., June 2005. Seeing stars: Exploiting class relationships for
  sentiment categorization with respect to rating scales. In: Proceedings of
  the 43rd Annual Meeting of the Association for Computational Linguistics
  (ACL'05). Association for Computational Linguistics, Ann Arbor, Michigan, pp.
  115--124.

\bibitem[{Polanyi and Zaenen(2006)}]{polanyi2006}
Polanyi, L., Zaenen, A., 2006. Contextual valence shifters. Computing attitude
  and affect in text: Theory and applications, 1--10.

\bibitem[{Socher et~al.(2010)Socher, Manning, and Ng}]{dlparsing}
Socher, R., Manning, C., Ng, A., 2010. Learning continuous phrase
  representations and syntactic parsing with recursive neural networks. In:
  Proceedings of the NIPS-2010 Deep Learning and Unsupervised Feature Learning
  Workshop.

\bibitem[{Socher et~al.(2011)Socher, Pennington, Huang, Ng, and
  Manning}]{socher2011}
Socher, R., Pennington, J., Huang, E.~H., Ng, A.~Y., Manning, C.~D., 2011.
  {Semi-Supervised Recursive Autoencoders for Predicting Sentiment
  Distributions}. In: Proceedings of the 2011 Conference on Empirical Methods
  in Natural Language Processing (EMNLP).

\bibitem[{Tu et~al.(2012)Tu, He, Foster, van Genabith, Liu, and Lin}]{subtrees}
Tu, Z., He, Y., Foster, J., van Genabith, J., Liu, Q., Lin, S., 2012.
  Identifying high-impact sub-structures for convolution kernels in
  document-level sentiment classification. In: Proceedings of the 50th Annual
  Meeting of the Association for Computational Linguistics: Short Papers -
  Volume 2. ACL '12. Association for Computational Linguistics, Stroudsburg,
  PA, USA, pp. 338--343.

\end{thebibliography}

\end{document}